# Stop Removing Stopwords: How an Inherited Preprocessing Default Distorts Legal Text-as-Data

**Gregory M. Dickinson**[1]

## Abstract

Empirical legal scholarship increasingly treats judicial text as data, and much of it still runs on sparse, interpretable pipelines—TF-IDF features and linear classifiers—because the textual feature is often the object of study rather than a means to a prediction. Yet these pipelines inherit preprocessing defaults from mid-century information retrieval that were never validated against classification accuracy. The most entrenched of these is stopword removal. This study introduces an exhaustive single-word ablation that measures a preprocessing step's effect directly against the downstream objective, and applies it to stopword removal as the hardest case to dislodge. Matching Supreme Court Database labels to Caselaw Access Project opinion texts, the study examines two binary tasks that bracket F1 headroom, ideological direction (no-removal baseline F1 ≈ 0.68) and constitutional versus non-constitutional law type (≈0.92), across 7,668 and 7,001 opinions. For each task, the ablation removes each of roughly 18,500 candidate words in turn, and a task-specific stoplist is built from the resulting measurements. Generic stoplists in common use fall below the no-removal baseline on held-out opinions in all twelve tests. The task-specific stoplists move held-out F1 by +0.0023 (95% CI [−0.0124, +0.0170]) on ideology and by +0.0001 ([−0.0082, +0.0085]) on law type. Neither task shows a detectable benefit from removal, and a supplemental analysis finds that word-level statistics predict a word's removal effect poorly, because the words' true removal effects differ by less than the measurement can register and so leave little to predict. The method generalizes to any inherited preprocessing default, and the result is a caution specific to interpretable legal text-as-data, where a step that reshapes which features a model sees can distort the doctrinal and ideological signal the research is meant to recover. The burden of proof sits with removal.

## 1. Introduction

As empirical legal scholars turn text into data, a methodological fork has opened. Transformer and large-language-model systems now dominate legal NLP benchmaks, yet a well-articulated constituency continues to favor sparse, interpretable models where the research aim is explanation rather than raw prediction (Livermore et al., 2024), where legal language is formulaic enough that lexical methods remain competitive (Han et al., 2026; Mori et al., 2025), and where the legal domain's demand for auditable, challengeable reasoning makes feature-level transparency a requirement (Kelsall et al., 2026; Medvedeva et al., 2023; Resck et al., 2025). For these scholars the operative features are part of the finding—the weight a model places on a phrase is itself evidence about judicial behavior (Truscott & Romano, 2025). That is precisely the setting in which an unexamined preprocessing default is most consequential. Preprocessing decisions of this kind are known to swing text-as-

[1]University of Nebraska College of Law, Lincoln, NE, USA

**Corresponding Author:**
Gregory M. Dickinson, University of Nebraska College of Law, 1875 N 42nd St, Lincoln, NE 68583-0902, USA.
Email: gdickinson@jd10.law.harvard.edu

data results in ways substantive theory does not anticipate (Denny & Spirling, 2018), and automated content analysis carries a standing obligation to validate such choices against the downstream task (Grimmer & Stewart, 2013). This article takes the most entrenched such default, stopword removal, and asks what it actually does to a legal classifier.

By convention, text-classification pipelines often begin by throwing words away. Before a model sees a document, a stoplist strips out *the*, *of*, *and*, and a few hundred other high-frequency words on the assumption that they carry no useful signal. The step is so routine that it is rarely questioned. Yet stoplists were engineered for mid-century information retrieval on two grounds that had nothing to do with predictive accuracy—Hans Peter Luhn's claim that the words most useful for telling documents apart fall in a middle frequency band, and the plain economy of a smaller index, which dropping common words shrank by 30 to 50% (Luhn, 1957, 1958; van Rijsbergen, 1979). The idea that stopword removal improves classification came afterward, as a retrospective justification for a habit that was already a default.

Belief about that habit runs along a gradient. The permissive default, kept alive by software settings and textbook treatments rather than by any peer-reviewed research, holds that a generic list is roughly free and more likely to help than to harm. A more informed skepticism, voiced, for example, in scikit-learn's documentation, warns that off-the-shelf English stoplists carry known problems and suit no task in particular, a caution this article shares and builds on. A third and more demanding position gives the practice its intellectual footing. That view holds that stoplists are worthwhile when built for a particular corpus and task by someone who knows the domain, and it is this view that has been asserted to justify building domain-specific stoplists, such as for biomedical research. That belief is rarely tested, however, because testing it is difficult and computationally expensive. For judicial text the question is wide open. Researchers who classify Supreme Court opinions make their stopword choices by convention, and no prior work has validated a corpus-specific stoplist for that setting.

This article tests the convention directly. It evaluates the generic lists researchers actually use, and it then makes a serious attempt to build the task-specific-type list that the demanding view calls for, scoring every candidate stopword, selecting the top scorers for removal, and then asking whether the stoplist thus produced improves classification on a holdout set of opinions. This study examines two binary classification tasks built from the leading U.S. Supreme Court Database. One task predicts the ideological direction of a decision and the other predicts whether or not it rests on constitutional law. The two tasks were chosen to sit at opposite ends of the difficulty range. On the ideology task, the no-removal baseline reaches an F1 of about 0.68, which leaves real room for a stoplist to help. On the law-type task, the no-removal baseline reaches an F1 of about 0.92, with little room left for improvement and thus a far stiffer test. A benefit from stopword removal anywhere across the range of difficulty should surface in at least one of the tasks.

For each task the analysis does two things. First, it evaluates classification improvement from application of three generic stoplists in common use. Second, it builds a task-specific stoplist by removing each of roughly 18,500 candidate words one at a time, refitting the entire pipeline after each deletion, and recording how that single removal moves the F1 score. The words whose removal helped by more than measurement noise alone would produce are then removed together, and the resulting list is scored on held-out opinions that played no part in its construction. A supplemental analysis asks, separately, whether the benefit of removing a word can be predicted from nine word-level statistics of the sort that implicitly lie behind an expert's intuition.

Three findings follow, all of which point against removal. First, the generic lists do not help. Across three lists, two tasks, and two settings of the feature pipeline, all twelve stopword removal tests leave the classifier no better off. The twelve tests share list members, corpora, and features, so they are not independent. But under every list, task, and specification tested, the point estimate moved toward harm and never toward benefit. Second, the task-specific lists do no better. Removing the stopwords selected by removal benefit shifts held-out F1 by +0.0023 (95% CI [−0.0124, +0.0170]) on the ideology task and by +0.0001 ([−0.0082, +0.0085]) on the law-type task. Both intervals include zero. The gains measured one word at a time do not combine for additive improvement when the words are removed together. Section 5 traces most of the shrinkage to a shared pool of borderline opinions that the stopwords flip in common. Third, the supplemental analysis finds that word-level statistics predict a word's removal effect poorly. Nine statistics that quantify the standard rationales for dropping a word, such as how common it is or how much its removal reshapes the surrounding phrases, explain little of the variation in measured removal effect. A test–retest analysis finds that the words' true removal effects differ from one another by less than the measurement can register, so there is little for any word-level statistic to predict. Still, whether some potential stopword list could ever help these classifiers remains an open question. No list this study tested, whether stock or purpose-built, delivered a benefit the design could detect. A step with no demonstrated benefit and a real cost (in added preprocessing and added researcher discretion) should have no claim to be the default. The sensible default is to stop removing stopwords.

This article makes three contributions. The first is methodological and general. Exhaustive ablation against the downstream metric is a reusable instrument for testing any inherited preprocessing convention—lowercasing, lemmatization, n-gram range, minimum document frequency—not stopword removal alone, and it replaces an expert's guess about which tokens to drop with a direct measurement of what dropping them does. The second is empirical. No published stoplist has been validated for judicial text, three lists in wide use fell below the no-removal baseline in every test, and a serious attempt to construct a task-specific list, on a task with ample headroom and on one with little, produced no improvement the design could detect. A researcher who reaches for a stoplist in this setting follows a convention with no demonstrated justification. The third is a caution specific to interpretable legal text-as-data. Because the feature is often the legal finding, a preprocessing step that silently rewrites the feature space can distort the signal a study means to measure. Removing the conjunctions in *cruel and unusual punishment* or *unreasonable search and seizure* mints the very trigrams a researcher wants, while the same step collapses the Constitution's *Privileges and Immunities Clause* and *Privileges or Immunities Clause* into one indistinct feature. What looks like neutral cleanup is a substantive choice about which doctrinal distinctions survive into the data.

The remainder of the paper develops these results in turn. It begins with the origins of stopword removal and the belief that sustains it, then describes the data and the pipeline, tests the generic lists, reports the exhaustive ablation and the held-out evaluation of the task-specific lists it produced, and closes with the implications for how legal text should be prepared for classification.

## 2. Background and Prior Work

### 2.1. The Information-Retrieval Origins of Stoplists

A stoplist is a roster of high-frequency words—*the*, *of*, *and*, and the like—that a text-processing pipeline discards before analysis. The device is old, and the two rationales that produced it had nothing to do with prediction. The first rationale was discriminative. Working at IBM in the late 1950s, Hans Peter Luhn proposed that a word's frequency tracks its significance, so that the terms most useful for telling one document from another fall in a middle frequency band, whereas very common and very rare words have low "resolving power" and make poor index terms (Luhn, 1957, 1958). The second rationale was economic. Discarding the most common words measurably shrank the index. Van Rijsbergen (1979) reports that removing them reduces the size of the document file by 30 to 50%. The stoplist thus combines linguistic discrimination with index economy. On neither account was the goal classifier accuracy.

The standard lists were built once and then frozen. Salton's SMART system gathered common words into a "negative dictionary," in what became the canonical stop list of early retrieval (Salton & McGill, 1983). Fox (1989) derived a 421-word list from frequency counts in the Brown Corpus—a one-million-word reference corpus of American English compiled in the 1960s—culling words too important to drop and adding others by hand. These artifacts then propagated, largely unexamined, into modern open-source software. Perhaps the most influential stoplist in contemporary natural language processing illustrates the pattern. The English list shipped with scikit-learn descends from one circulated by the information-retrieval group at the University of Glasgow, van Rijsbergen's own institutional home. Its provenance went unexamined for years, and Nothman et al. (2018) report that scikit-learn's maintainers were long "unaware of how it was constructed." These lists are frozen inheritances, not task-optimized instruments.

Importantly, the very field that invented stoplists has since abandoned them. As retrieval systems acquired good index compression and inverse-document-frequency weighting, the storage savings evaporated, and practice moved from large stop lists of two to three hundred terms, to very small ones, to—in the words of Manning et al. (2008, §2.2.2)—"no stop list whatsoever." Stopword removal no longer survives in the field that produced it. It persists in the one that borrowed it—in natural language processing.

That migration left a conceptual seam that this article takes up. Lists developed to address concerns about which terms discriminate in retrieval, or about how large an index grows, carry no automatic implication about whether removing those terms helps a classifier. When the technique crossed into predictive NLP, no one tested that gap.

### 2.2. The Migration Into NLP and the Anatomy of the Live Belief

Stopword removal became a fixture of text classification accidentally, by inheritance. The field adopted the technique based on its information-retrieval connection and never revalidated it for prediction. It is taught as routine preprocessing (Jurafsky & Martin, 2026; Manning et al., 2008), shipped as a single keyword in the most widely used feature-extraction library—scikit-learn's stop_words = 'english'—and expected by reviewers. Nothman et al. (2018) trace the technique's spread to open-source resources, which, they observe, "tend to become de-facto standards by virtue of their availability and popular use," and once a list ships in a library it is applied without scrutiny.

Two accuracy rationales were supplied after the fact. The first is low-information pruning. Function words are presumed uninformative about a document's class, so dropping them is said to concentrate the model on content and raise the signal-to-noise ratio (Sarica & Luo, 2021). The second is trigram bridging. Deleting an intervening function word lets the vectorizer treat the surrounding content words as adjacent, minting collocations the raw n-gram space would miss. Of the two, bridging is the stronger rationale and the more particular to classification; Section 2.3 develops it in full.

Current belief falls into three categories. The working default is a permissive view, according to which generic stopword lists are roughly free—likely to help, unlikely to hurt. Although the peer-reviewed literature does not generally advocate stopword removal, the habit lives in library defaults and textbook treatments, which is what makes it durable. The same default operates in legal text-as-data work. For example, a study predicting justices' votes from oral-argument transcripts stripped stopwords before extracting n-gram features on the ground that such words "generally decrease the effectiveness of n-gram feature analysis," an assertion offered without citation or test (Dickinson, 2019, p. 458). An informed skepticism opposes that default, suggesting that generic stoplists are a mistake for serious work. scikit-learn's own documentation cautions that its built-in english list has "several known issues" and is no one-size-fits-all solution, since popular lists routinely include words—*computer* is the stock example—that are highly informative for some tasks. This article shares that skepticism and builds on it. A third, more nuanced, position holds that domain-specific stopword lists, unlike generic ones, can improve classification performance when built with sufficient subject-matter expertise. This belief supposedly justifies constructing special legal or biomedical stoplists, yet almost no one has tested it.

Researchers have built domain-specific lists by frequency threshold (Fox, 1989), by similarity statistics in biomedical text (Wilbur & Sirotkin, 1992), by collection-specific sampling on TREC data (Lo et al., 2005), by supervised extraction from labeled documents (Makrehchi & Kamel, 2008), by entropy measures in topic models (Gerlach et al., 2019), and by term frequency and entropy on patents (Sarica & Luo, 2021). The conclusion to date has been that domain lists sometimes help a little and sometimes do not, under conditions too heterogeneous to settle the matter. Gerlach et al. (2019) observe that manual lists "cannot be readily generalized across knowledge domains or languages."

But the empirical record supporting even domain-specific stoplists is thinner than the practice's confidence implies. The clearest negatives come from topic modeling, where removal's apparent benefit is largely superficial (Schofield et al., 2017), and from Twitter sentiment, where precompiled lists degrade performance (Saif et al., 2014). Head-to-head tests in supervised document classification, the setting that matters here, are rare (Sarica & Luo, 2021; Silva & Ribeiro, 2003). For legal text, the record is empty. No prior peer-reviewed work proposes or validates a corpus-specific stoplist for judicial opinions.

This article fills that gap by a different route. Every prior method guesses at removal candidates from word-level properties—frequency, entropy, similarity—and trusts that the removals will help. Instead of guessing, the analysis measures each stopword removal candidate's marginal effect on classification accuracy directly, one word at a time. Done exhaustively, this amounts to a serious attempt to build a domain-specific stoplist, and the results suggest the limits of what such an attempt can accomplish.

## 2.3. The N-gram Bridging Mechanism

Of the two accuracy rationales in Section 2.2, bridging is the more plausible. Its force comes from the order in which a vectorizer operates. scikit-learn applies any stoplist to the token stream before n-grams are formed, so that formerly separated words flanking a removed token come together and the vectorizer mints a new n-gram that the raw text does not (Pedregosa et al., 2011). Consider the phrase *Secretary of the Treasury*. Left intact, it produces the bigrams *Secretary of* and *the Treasury*, but never the pair a human reader would recognize as the unit of meaning. Strip *of* and *the* first, however, and the bigram *Secretary Treasury* appears. The content-word skeleton survives as a single feature that fires whenever the phrase is used, in place of two uninformative fragments.

Bridging is a form of feature engineering. Legal writing is thick with multiword terms whose two content words sit a preposition and an article apart, as in *freedom of the press*, *preponderance of the evidence*, and *breach of the peace* (Justeson & Katz, 1995). A span that wide is too long for a trigram model to reach, so the two content words never share a feature unless the function words between them are removed or the n-gram range is widened to four-grams, a setting few analysts would choose. The pairs at stake are collocations, words that recur together far more often than chance would predict and carry meaning as a unit (Church & Hanks, 1990), and stopword removal is the cheap way to make them visible to an ordinary contiguous n-gram pipeline.

But the mechanism cuts both ways. Every removal that surfaces useful collocations will also sever legitimate collocations that contained the removed word. Deleting a token erases its own unigram and dissolves each bigram and trigram that included it. Intuition regarding what words are "valuable" is unreliable here, because a word can be a productive bridge in one phrase and indispensable content in the next. Whether the trade comes out positive depends on the balance of bridged gains against severed losses across the whole vocabulary, a balance set by the task and the corpus as a whole. No single word's properties can settle it. There is no theoretical reason to expect that balance to favor removal.

The mechanism does, however, leave testable traces. Collocations spanning longer than the specified n-gram range become visible to a vectorizer only when an internal stopword is removed. So a removal effect that differs between a range extending to trigrams (1,3) and one capped at bigrams (1,2) is evidence that bridging is doing something (Wang & Manning, 2012). This article uses that contrast throughout as a diagnostic of whether the predicted bridging mechanism has an effect. The benefit or detriment of that effect is a separate question, and one that a simple contrast cannot settle. As Sections 4 and 5 show, the two ranges behave differently on the two tasks, which confirms that bridging is active. Yet the difference does not resolve into a reliable advantage on held-out data.

## 2.4. SCOTUS Classification: SCDB, CAP, and Standard Pipelines

Quantitative study of the Supreme Court rests on a common foundation, the Supreme Court Database (SCDB) (Spaeth et al., 2025), which codes the Court's decisions reaching back to the Court's earliest terms; the analysis draws on the modern era, from 1946 onward. Two hand-coded variables come from SCDB, decisionDirection and lawType. The first task uses the decisionDirection variable, which records the ideological direction (liberal or conservative) of each decision. The second uses the lawType variable, collapsed into a binary contrast between cases decided on constitutional grounds and all others. The SCDB supplies the variable codings but no opinion text, so the text of the decisions is retrieved from the Caselaw Access Project (2024) and each coded decision is matched with its associated text. Section 3 describes the matching and filtering.

Both decisionDirection and lawType sit within a mature literature on measuring and predicting judicial behavior. One long tradition estimates justices' ideology from their votes, most influentially through the dynamic ideal-point model of Martin and Quinn (2002); a parallel tradition predicts case outcomes from the SCDB's own structured variables (Katz et al., 2017). Automated content analysis brings these questions to opinion text itself (Evans et al., 2007), and a growing body of work now infers ideology and related signals directly from the Court's writing. Researchers have classified the ideological direction of judicial opinions (Hausladen et al., 2020), sorted opinions by issue area (Chalkidis et al., 2022; Undavia et al., 2018), recovered justice preferences from briefs (Feldman & Menounou, 2015), scaled opinions jointly with votes (Lauderdale & Clark, 2014), modeled the thematic content of appellate opinions (Livermore et al., 2024), and detected ideological stance in the Court's language (Bergam et al., 2022; Truscott & Romano, 2025). Prior work thus establishes the premise of this study, that the text of an opinion carries recoverable ideological and doctrinal signal.

Across text-based studies the workhorse models are consistent. Sparse bag-of-n-grams features, weighted by TF-IDF and passed to a linear classifier such as logistic regression or a support vector machine, remain the standard and most interpretable approach to legal text classification. They remain competitive with and at times surpass the performance of heavier neural models (Aletras et al., 2016; Chen et al., 2022; Soh et al., 2019; Şulea et al., 2017). Transformer models have entered the field as well (Chalkidis et al., 2020, 2022), but they split text into subwords and apply no stopword list at all, so the question studied here never arises for them. Moreover, their predictions are difficult to interpret, which limits their usefulness for the explanatory aims of much empirical legal research, where knowing which features drive a classification is frequently more important than the classification itself. The tests therefore run inside the sparse n-gram pipeline, the framework most widely used for this task and the one where bridging would be expected to pay off, if anywhere at all.

# 3. Data and Pipeline

## 3.1. Data Sources

The two datasets share a common construction. Case labels come from the 2025 release of the SCDB (Spaeth et al., 2025), taking decisionDirection for the ideology task and lawType for the law-type task. The database codes decisions without supplying their text, so each majority opinion is retrieved from CAP (2024), matching on U.S. Reports citation where available and falling back to the Supreme Court Reporter citation and then to the docket number. Three filters follow, dropping cases whose opinion text cannot be located, cases whose lawType resists the constitutional/non-constitutional collapse (e.g., common law and original jurisdiction decisions), and opinions shorter than 2,000 characters, which are procedural stubs carrying no substantive reasoning. A stratified ten-percent holdout is then reserved for each task, balanced on label and term decade. The final ideology corpus contains 7,668 opinions (6,902 training and 766 holdout) and the law-type corpus 7,001 (6,301 training and 700 holdout).

## 3.2. Two Classification Tasks

This study examines two binary classification tasks, chosen to sit at opposite ends of the baseline-accuracy range. The first predicts the ideological direction of a decision, the liberal or conservative coding supplied by decisionDirection. On this task, the baseline pipeline reaches mid-range accuracy (F1 $\approx$ 0.68), leaving substantial headroom for a preprocessing step to improve upon. The second task predicts whether a case was decided on constitutional grounds, the collapsed form of lawType. Here the same pipeline already scores high (F1 $\approx$ 0.92), leaving little room to rise. Section 3.4 reports the baseline figures for both tasks.

The pairing is deliberate, since stopword removal should pay off, if anywhere, where accuracy is unsettled and collocations remain to be recovered, which makes the ideology task the friendliest case for the bridging mechanism of Section 2.3. The law-type task is the stress test, the harder case for a gain. Together the two tasks bracket the range of headroom in which a benefit of stopword removal could appear, so a gain anywhere in that range should surface on one task or the other. Section 5 reports the result.

## 3.3. The Baseline Pipeline

The pipeline is deliberately conventional, the basic n-gram architecture described in Section 2.4. Each opinion is vectorized with TF-IDF over the (1,3) n-gram range, wide enough for the bridging mechanism of Section 2.3 to operate, yet narrow enough to keep the feature space manageable. The vocabulary is capped at the 150,000 most frequent features, which keeps the roughly 18,500 ablation retrains of Section 5 computationally feasible and costs little accuracy relative to the uncapped vocabulary. Term frequencies are scaled sublinearly, as log (1 + tf), so that the longest opinions—some run past 350,000 characters—do not swamp the shortest through length alone.

The classifier is an L2-regularized logistic regression (Pedregosa et al., 2011). The inverse-regularization strength C was grid-searched across six values spanning 0.01 to 1,000 under five-fold cross-validation. Cross-validated F1 climbs and then flattens, peaking at C = 100 on both tasks, with C = 10 trailing by roughly 0.001 on ideology and 0.003 on law type (Table S1). C = 10 is fixed for both tasks. A single shared value keeps the two experiments directly comparable, and because the gap between C = 10 and C = 100 is smaller than the variation across folds, the smaller C, which applies the stronger penalty, is the conservative choice.

Logistic regression is preferable to the other standard choices for reasons specific to the ablation. It refits faster than a support vector machine across the thousands of ablation runs, and its coefficients are signed and directly interpretable, unlike those of tree ensembles. Logistic regression is also the workhorse of prior legal-text classification, which keeps the results comparable to that literature.

The pipeline stays fixed throughout, and only the set of words removed before vectorization varies. The no-removal condition is the reference, and every reported delta measures a removal condition against it.

### 3.4. Baseline Performance

Using the pipeline of Section 3.3, ideology reaches a held-out weighted F1 of 0.6819 and law type 0.9185, a gap of roughly 0.24 (Table 1). The two corpora differ in balance. Ideology divides almost evenly between liberal and conservative decisions, whereas law type leans toward non-constitutional ones.

## 4. Do Standard Stoplists Help?

Three generic English stoplists are in common use—scikit-learn's ENGLISH_STOP_WORDS (318 words), NLTK's English list (198 words), and the list bundled with spaCy's en_core_web_sm model (326 words). Each is tested by removing its words from every opinion before vectorization and retraining the full pipeline on what remains. Each list's effect is reported as the change in holdout weighted F1 against the no-removal baseline, and the experiment runs twice, once for the (1,3) n-gram range and once for the (1,2) range. Table 2 collects the results.

Across three lists, two tasks, and two n-gram ranges, stopword-removed opinions classified below the baseline in all twelve tests. The harm is larger on the ideology task, where the holdout deltas run from −0.0048 to −0.0168, and smaller on the law-type task, where they run from roughly −0.0013 to −0.0057. A baseline that already scores this high leaves less room for a list to do harm. Each delta is small enough that its bootstrap confidence interval includes zero (Section 6.3 returns to the question of power), so none of the harms caused by stopword removal is statistically significant at these holdout sizes. The deltas' consistent negative direction is still informative, though it carries no statistical weight of its own. The twelve tests are correlated rather than independent. The three lists have much of their membership in common, the two tasks are built from largely the same opinions, and the two n-gram ranges draw on the same unigrams and bigrams, so the tests overlap at every level of the design. No formal probability attaches to the pattern, because correlated tests do not compound evidence the way independent ones would. Under every list, every task, and every specification tested, the point estimate moved toward harm and never toward benefit, and the likeliest explanation is that removing several hundred words at once dissolves more useful collocations than it bridges new ones. List length does not track the damage. NLTK at 198 words and spaCy at 326 land close together on ideology, so a longer list does not simply do proportionally more harm. Damage tracks which words a list removes and which collocations that removal disrupts. The results do not show that generic stoplists harm the classification of judicial opinions. They do show that a practice adopted on the assumption that it helps produced no estimate in that direction here. A researcher who removes these words is relying on a convention that no test in this study supports. The burden of showing that removal helps therefore lies with the researcher who removes, and nothing in these tests meets it.

Re-running each condition under the (1,2) range tests whether trigram-scale bridging contributes to the removal

**Table 1.** Baseline Performance of the No-Removal (1,3) Pipeline of Section 3.3 at C = 10, Under Five-Fold Cross-Validation on the Training Set and on the Held-Out Opinions (*N* = 766 for Ideology, *N* = 700 for Law Type)

| Task | Class balance | Five-fold CV F1 | Holdout F1 |
|---|---|---|---|
| Ideology | 51.3% liberal, 48.7% conservative | 0.6531 | 0.6819 |
| Law type | 36.9% constitutional, 63.1% non-constitutional | 0.9116 | 0.9185 |

All values are weighted F1; the stratified holdout preserves each task's corpus class balance.

**Table 2.** Generic Stoplists Evaluated Against the No-Removal Baseline, Under Both n-gram Ranges. List Size is the Number of Words in Each Published List. Holdout F1 and Δ are Computed on the Held-Out Opinions (Ideology *N* = 766, Law Type *N* = 700)

| N-Gram range | Condition | List size | Ideology holdout F1 | Ideology Δ | Law-type holdout F1 | Law-type Δ |
|---|---|---|---|---|---|---|
| (1,3) | Baseline | 0 | 0.6819 | — | 0.9185 | — |
| (1,3) | sklearn | 318 | 0.6676 | −0.0143 | 0.9171 | −0.0013 |
| (1,3) | NLTK | 198 | 0.6771 | −0.0048 | 0.9172 | −0.0013 |
| (1,3) | spaCy | 326 | 0.6651 | −0.0168 | 0.9157 | −0.0028 |
| (1,2) | Baseline | 0 | 0.6897 | — | 0.9185 | — |
| (1,2) | sklearn | 318 | 0.6766 | −0.0131 | 0.9128 | −0.0057 |
| (1,2) | NLTK | 198 | 0.6778 | −0.0118 | 0.9171 | −0.0014 |
| (1,2) | spaCy | 326 | 0.6738 | −0.0159 | 0.9129 | −0.0056 |

Every holdout Δ is negative, and each individual delta's bootstrap confidence interval includes zero.

effect. A collocation that spans a trigram becomes visible to the vectorizer only when an internal word is removed, so a removal effect that shifts between (1,3) and (1,2) is a sign that the bridging mechanism of Section 2.3 is at work. The generic lists are too blunt to isolate it; because they bundle hundreds of removals together, they stay negative under both ranges on both tasks. The mechanism shows itself instead in the single-word removal tests that Section 5 develops. The removal lists those tests produce have holdout deltas that move from +0.0023 under (1,3) to −0.0043 under (1,2) on ideology, a swing of 0.0066, and from +0.0001 to −0.0057 on law type, a swing of 0.0058—nearly the same distance (Section 5 reports these figures in full, with confidence intervals). The (1,3)-versus-(1,2) comparison confirms that bridging operates, but does not show that it improves classification performance. On neither task does the (1,3) delta reach a positive estimate distinguishable from zero. Each bridge a removal opens is paid for by a collocation it severs elsewhere, so even though the mechanism is plainly active, it produces no reliable gain on held-out opinions. Section 5 takes up the question this leaves open: whether a list built for these tasks by testing every candidate word one at a time does better than leaving the text alone. Such a list produces no detectable improvement on either task.

## 5. Can a Task-specific Stoplist Do Better?

### 5.1. The Ablation Design

The ablation design tests each candidate word's removal directly. For each candidate, every occurrence is deleted from the training opinions, the full (1,3) pipeline is refit, and the change in weighted F1 against the no-removal baseline is recorded as that word's ΔF1. The delta measures a removal's benefit against the objective stopword removal is claimed to serve, rather than inferring it from the word's frequency or entropy. The candidate pool is every unigram in the baseline vocabulary of Section 3.3, that is, every single word that appears in at least five opinions and ranks among the 150,000 most frequent features: 18,575 words for ideology and 18,557 for law type. Every candidate is tested; the words whose deltas clear the noise floor described below form each task's removal list. That list is what a serious single-word search for a task-specific stoplist produces, and Section 5.3 reports how it performs on held-out opinions. Runtime, worker allocation, and the selection rule are reported in Supplemental Section B.

Each ΔF1 is measured under three-fold cross-validation. The training opinions are split once into three folds, and the same split is used for the baseline and for every ablated pipeline, so that each ΔF1 compares the ablated pipeline and the baseline on identical folds. Because the F1 deltas are the difference between two cross-validated scores—classification F1 with and without the candidate stopword—each carries measurement error, and a word whose measured benefit is smaller than that error cannot be distinguished from a word whose removal does nothing. The noise floor screens such words out. The error scale is measured directly. Rerunning the ablation for 1,150 ideology words and 1,131 law-type words, among them the three hundred highest-document-frequency words, under a second assignment of opinions to folds remeasures the same deltas. Comparing the two runs for those three hundred words separates the stable part of the deltas from the run-to-run error and gives the typical size of that error, a standard deviation across candidates of 0.000948 on ideology and 0.000325 on law type (Supplemental Section B.3 gives the decomposition). The floor is set above that scale, at 0.001145 on ideology and 0.000354 on law type, so that a candidate enters the removal list only when its measured benefit exceeds that error. As a sensitivity check, Supplemental Section B.1 reports the lists that a stricter floor of twice the measured error would admit, and evaluates them on the holdout. Every interval again includes zero, as in Section 5.3.

### 5.2. The ΔF1 Landscape and Its Scarcity

The exhaustive ablation returns a landscape that is flat and faintly negative. Across all candidates, ΔF1 averages −0.000396 on ideology and −0.000128 on law type. Law type's distribution is the tighter of the two, but both share a shape, peaked at zero and weighted toward harm (Figure 1). Only 24.5% of ideology candidates and 21.9% on law type raise F1 at all, most of them imperceptibly. Removing the typical word leaves F1 unchanged, and the removals that do move F1 hurt more often than they help.

Applying the Section 5.1 noise floor thins the beneficial removals further. Only 78 ideology removal candidates clear it and only 95 for law type, 0.42 and 0.51% of the respective pools, or about one word in two hundred. An exhaustive single-word search over more than eighteen thousand words surfaces at most a few dozen candidates worth removing. Those candidates form the removal lists that Section 5.3 evaluates on held-out opinions.

## *5.3. Holdout Evaluation of the Task-specific Stoplists*

The removal lists of Section 5.2, 78 words for ideology and 95 for law type, are evaluated on held-out opinions. The pipeline is retrained with each task's list deleted before vectorization, scored on the holdout, and the change in weighted F1 against the no-removal baseline is reported with paired bootstrap confidence intervals (Table 3). On ideology, the list moves holdout F1 by +0.0023 (95% CI [−0.0124, +0.0170]) over 766 opinions. On law type, it moves holdout F1 by +0.0001 (95% CI [−0.0082, +0.0085]) over 700 opinions. Under the (1,2) range the same lists move holdout F1 by −0.0043 (95% CI [−0.0174, +0.0088]) on ideology and −0.0057 (95% CI [−0.0157, +0.0042]) on law type, the swing noted in Section 4. Every interval includes zero. The study does not detect an effect from either list under either range. The ideology (1,3) interval runs from a harm of 0.0124 to a benefit of 0.0170, which shows how large an effect could go undetected on a holdout of this size.

The F1 gains measured on the holdout set, +0.0023 and +0.0001, are small beside the sums of the single-word gains from the selected words, +0.1066 on ideology and +0.0568 on law type. Most of that difference is because many of the gains in each list are attributable to words whose removals shift the classifications of the same opinions, so the sum repeatedly counts them. To show that, all 78 ideology words were removed together and rescored on the same three folds that produced the single-word estimates. The joint removal yields a gain of +0.0083 where the single-word gains sum to +0.1066, and that shortfall accounts for 94.2% of the difference between the sum and the holdout gain; on law type the joint removal yields +0.0026 where the single-word gains sum to +0.0568, and the shortfall accounts for 95.6% (Supplemental Section B.2, Table S3). A word's ΔF1 is, in effect, a count of the training opinions whose classification its removal changes, gains less losses, and on ideology the 78 single-word removals change a classification 2,890 times between them while only 330 distinct opinions are ever involved, 8.76 removals per opinion; the law-type ratio is 8.71. An opinion counted once for every word whose removal changes its classification in the sum is counted at most once when the words are removed together (Supplemental Section B.2). The remaining 5.8 and 4.4% is lost between the

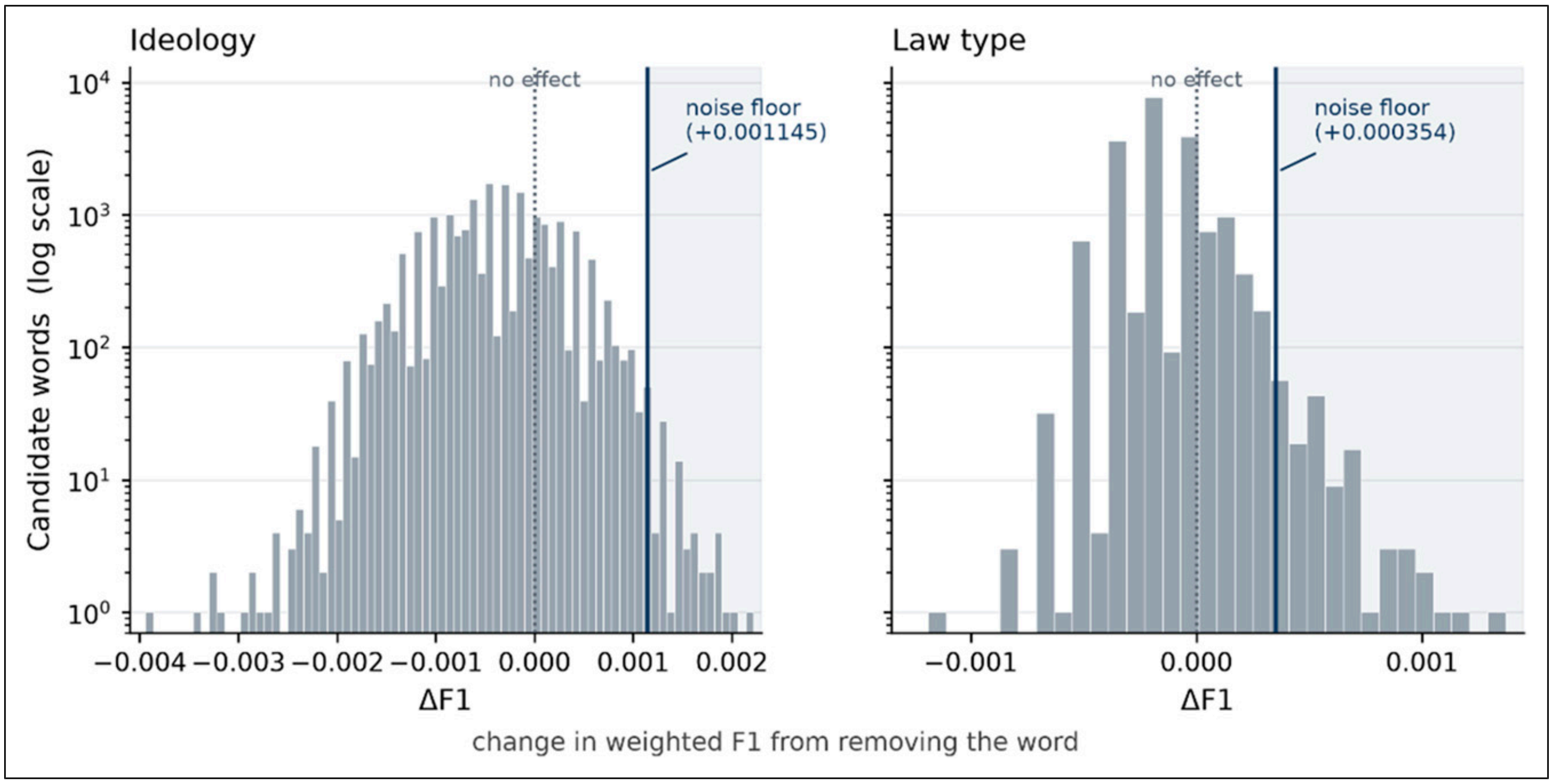

**Figure 1.** Distribution of single-word ablation ΔF1 across candidate words, for ideology (18,575 candidates) and law type (18,557), on a log count scale. Negative values indicate that removing the word lowers weighted F1. Both distributions concentrate near zero with a slight leftward weight, and on each task the noise floor of Section 5.1 (0.001145 on ideology, 0.000354 on law type) sits in a thin positive tail that few words reach.

**Table 3.** The Removal List of Each Task Evaluated Against the No-Removal Baseline Under the (1,3) n-gram Range, With Paired Bootstrap 95% Confidence Intervals on the Holdout Δ (10,000 Resamples). Holdout F1 and Δ are Computed on the Held-Out Opinions (Ideology $N$ = 766, Law Type $N$ = 700)

| Condition | Ideology holdout F1 | Ideology Δ [95% CI] | Law-type holdout F1 | Law-type Δ [95% CI] |
|---|---|---|---|---|
| Baseline | 0.6819 | — | 0.9185 | — |
| Removal list | 0.6842 | +0.0023 [−0.0124, +0.0170] | 0.9186 | +0.0001 [−0.0082, +0.0085] |

The ablation removal list contains 78 words for ideology and 95 for law type.

joint removal scored under cross-validation and the same removal scored on the holdout, that is the part of the cross-validated gain that does not carry over to unseen opinions (Supplemental Section B.2).

A reader who takes stopword removal of some kind as given, the field's default and something reviewers expect to see, might ask only which list to use rather than whether to remove at all. On the point estimates the removal lists come out ahead of every generic list on ideology, by as much as +0.0191 against spaCy, which would seem to settle that narrower question in favor of a task-specific list. The comparison does not detect that margin either. Each removal list is scored against each generic list on the same resampled holdout opinions, so that the sampling noise the two share cancels, and every one of the twelve paired intervals includes zero, on both tasks and under both n-gram ranges (Table S5; Supplemental Section C explains why the paired design leaves margins of this size unresolved). The question this article puts is the prior one, whether removal should be performed at all, and its recommendation on that question (Section 6.1) rests on the holdout result above and on the generic-list results of Section 4 rather than on the margin between them.

The two tasks were chosen to bracket the range of headroom in which a benefit of removal could appear (Section 3.2). Neither the task with the most room to improve nor the one with the least shows a detectable improvement from the list this search produced.

### 5.4. *A Helper Mechanism that Also Hurts*

Section 5.3 reports that the task-specific lists show no detectable improvement on held-out opinions over removing nothing at all. This section explains that result by tracing the removal mechanism through a few of the Court's stock phrases. The short answer is that stopword removal is too blunt an instrument. The same removal that builds a useful feature in one phrase destroys another in the next, for the stoplist, which acts on a word in every place it appears, cannot tell the two apart.

The bridging mechanism of Section 2.3 operates in this corpus. Consider the conjunction *and*, whose removal raises ideology F1 by 0.0019, enough to clear the noise floor and earn a place on the removal list. Many of the Court's stock constitutional phrases run four words long and set *and* at the hinge—*keep and bear arms*, *unreasonable search and seizure*, *cruel and unusual punishment*. A four-word span is too wide for any trigram to connect the content words, so the no-removal baseline models never register them as a unit. Removing the conjunction brings the content words close enough for a trigram to join them, so the vectorizer sees *keep bear arms*, *unreasonable search seizure*, and *cruel unusual punishment* as features. This is the bridging that stopword removal is supposed to produce.

Yet the same stroke that opens these bridges closes others. Removing *and* also dissolves every collocation that contained it, a cost that is easy to overlook because it is spread thinly across the vocabulary. A comparison of common conjunctions helps illustrate the point. Removing *and* helps by 0.0019, while removing *or* hurts by almost exactly as much, −0.0019, even though the two words share a frequency, a part of speech, and a place on every generic list. Removal effect is unpredictable even for words this similar, because the outcome turns on whether the trigrams a removal creates are worth more than the collocations it destroys. Even such simple words as *and* and *or* contain real information. Consider the Constitution's *Privileges and Immunities Clause* and its *Privileges or Immunities Clause*. Standard stoplists that remove the conjunctions *and* and *or* collapse the two, leaving only the less informative *privileges immunities*.

The net effect of any removal is the sum of its gains and losses. Across the full vocabulary, this sum runs negative for generic lists. For the task-specific lists the holdout evaluation does not detect a difference between removal and no removal. Whether *and* should be removed depends on the phrase it sits in, but stoplists act only word by word; they either strip a word everywhere or leave it everywhere.

## 6. Discussion

### 6.1. *Practical Guidance*

For classifying Supreme Court opinions with a TF-IDF pipeline, the right default is to remove no stopwords at all. The recommendation rests on comparing the costs and prospective benefits of each choice, not on a claim that removal can *never* help. Leaving stopwords in place forgoes, at most, a benefit the holdout evaluation could not detect. The task-specific lists that the single-word search of Section 5 produced moved held-out F1 by +0.0023 on ideology and +0.0001 on law type, and every confidence interval includes zero, on both tasks and under both n-gram ranges. The widest of those intervals, on ideology, admits a benefit as large as 0.0170, so a gain of that size could sit undetected on a holdout of 766 opinions. That figure is not an estimate of the benefit; it is the largest benefit the study fails to rule out. The benefit the holdout actually measured, +0.0023, is close to zero, and the generic lists that researchers do use moved every point estimate toward harm (Section 4).

Removing stopwords, by contrast, carries costs that are certain. It adds a preprocessing step, and it adds a discretionary choice among lists and thresholds that the no-removal choice does not require. A third cost is specific to interpretable legal text-as-data. Because the weighted feature is itself the object of study, a removal that reshapes the feature space can shift the doctrinal or ideological quantity a study means to recover, so leaving stopwords in place is also a matter of measurement validity. A benefit that a holdout of this size cannot separate from zero does not justify those costs. The default therefore should be no removal, and the burden of showing that a stoplist helps rests with the researcher who applies one.

Generic lists are the worst-supported option. The point estimate fell below the no-removal baseline in all twelve generic-list tests, on both tasks and under both n-gram ranges (Section 4). The likeliest explanation is that stripping several hundred connective words at once severs more useful collocations than it bridges. Section 2.2 already noted the informed skepticism about off-the-shelf lists, and no test here gives a reason to set that skepticism aside. A researcher who applies one of these lists to judicial text is following a convention that nothing here supports.

A purpose-built list does not change the recommendation. The task-specific lists of Section 5 came out of a serious attempt to build a beneficial stoplist. That attempt covered roughly 18,500 candidate words per task and consumed more than four hundred hours of computation across two ablation runs (Supplemental Section B). It produced no improvement on held-out opinions that the design could detect. Whether some other list could help these classifiers remains open. No published stoplist has been validated for judicial text, and no list this study tested, stock or purpose-built, delivered a detectable benefit. Unless and until a stoplist of either kind shows a benefit, the default should be to remove nothing.

These conclusions rest on a TF-IDF and logistic-regression classifier applied to SCOTUS opinions, and Section 6.3 takes up how far they travel.

## 6.2. Beyond Stopwords: Other Preprocessing Defaults

Both the generic and task-specific stoplists tested in this study were built using word-level criteria. Generic lists like those of Section 4 select words by word frequency, grammatical class, and linguistic intuitions about words' importance in general English. These generic lists fell below the no-removal baseline in all twelve tests. The task-specific lists of Section 5 similarly select by a word-level criterion, each word's own measured effect on this classifier and this corpus. Those lists moved held-out F1 by amounts whose intervals include zero (Section 5.3). The decomposition of Section 5.3 shows what limits the benefits even of task-specific lists. The words whose removal helped one at a time act largely on the same borderline opinions. Whether removing a word helps is fixed by the collocations it sits inside and the features its deletion makes reachable (Section 2.3). Those questions are settled only once the surrounding vocabulary, the corpus, and the classification task are all in place. A generic stoplist criterion that scores a word by frequency or grammatical class cannot account for that setting. The task-specific stoplists were created by scoring each word by its own single-word ablation effect on F1, so they do account for the corpus and classification task—but only on an all-or-nothing basis—remove the word in all occurrences or in none.

The caution from this study generalizes beyond stopwords to any word-level preprocessing decision. Several routine steps are decided word by word but, like stopword removal, reshape the n-gram space that downstream models work on (Section 2.3). Stemming and lemmatization fold related word forms into one token, collapsing phrases the vectorizer would otherwise keep distinct. A minimum-document-frequency cutoff discards rare tokens and the multiword features built on them. Each choice is made on a word's own statistics but plays out in the n-gram features it leaves standing. Whether collapsing *argued* and *arguing* into one stem helps, for example, turns on whether the phrases it merges were pushing the classifier the same way or opposite ways, the same question that neither a generic list nor a single-word search answered for stopwords. The same test is available for each of these steps, an ablation that measures the step directly against the downstream objective and scores the result on held-out opinions. No such test has been run here. Until one is, a stemmer or a frequency cutoff carries the burden Section 6.1 places on stopword removal: to show a benefit on held-out opinions before it is applied.

A supplemental analysis is consistent with this reading, approached from the other direction. Supplemental Section D asks whether a word's measured ΔF1 can be predicted from nine word-level statistics that formalize the usual reasons for removing a word: how common it is and how evenly it spreads across opinions, how varied its neighbors are, how much its presence reveals about an opinion's label, how heavily the baseline classifier already leans on it, and how strongly its removal would fuse the words around it into new collocations. Evaluated on words held out from the models' training data, the nine statistics explain almost none of the variation in ΔF1—$R^2$ of −0.007 on ideology and 0.038 on law type. A classifier trained to separate the most helpful removals from the most harmful ones scores an AUC of 0.50 and 0.54. A fit this poor admits two possible readings. The statistics may carry no usable signal, or the ΔF1 values they are asked to predict may be so largely measurement error that too little real variation remains for anything to predict. Supplemental Section D says which applies, drawing on the test–retest decomposition of Supplemental Section B.3. The second reading applies on both tasks, and in a particular form. The words' true removal effects differ from one another by less than the measurement registers, so a poor fit means that there is little for word statistics to predict, not that the statistics are uninformative. Neither reading affects the holdout results above, which measure stopword removal's benefit directly rather than asking what predicts it.

### *6.3. Limitations*

These conclusions are subject to four limitations. First, the ablation scores each word's removal individually and then builds from words that clear the noise floor. Because words are selected for the stoplists based on the effects of their individual removal, but the resultant lists are scored based on the words' removal collectively, the design cannot find a combination of words whose joint removal changes an opinion's classification when no member of the combination changes it alone. No search over such combination removals was run. Second, the holdout sets are small, 766 opinions for ideology and 700 for law type, so no single delta separates cleanly from zero. The confidence intervals of Table 3 span roughly ±0.015 F1 on ideology and ±0.008 on law type, so an effect smaller than that, whether a benefit or a harm, would go undetected. Section 6.1 weighs a benefit of that size against the costs of removal.

Third, only one pipeline is tested, TF-IDF features with logistic regression. Support vector machines, naive Bayes classifiers, and tree ensembles may behave differently. The bridging mechanism has its widest scope in sparse n-gram models, however, so the representation tested here is the one most hospitable to removal. Fourth, the study covers two binary tasks on a single corpus of Supreme Court majority opinions, a distinctive genre. Multiclass tasks with more headroom, and other legal genres such as legislative and regulatory text, remain untested.

### *6.4. Future Directions*

Three extensions would test whether the result generalizes. A first concerns combinations of words. Single-word ablation scores each removal on its own and so cannot select for a gain that appears only when words are removed together, and no test of that kind was run here (Section 6.3). For lists built one word at a time, the decomposition of Section 5.3 accounts for the gap between the sum of the single-word gains and the gain from removing the selected words together; a search that removes words in combination is the one selection regime that decomposition does not test. Probing pairs, or working through larger subsets, would measure what improvements joint removal might yield. A second extension would move to a multiclass task, such as the fourteen issue areas in the Supreme Court Database, where baseline classification performance is likely to be lower, leaving more room for stopword removal to help. A third would replicate the analysis elsewhere. Legislative and regulatory corpora, like judicial opinions, are dense with content words bound by connectives, and so are the natural next tests.

## 7. Conclusions

This article asked whether stopword removal, inherited from information retrieval and retroactively justified as a way to drop low-information words and bridge the content words they separate into new n-gram features, improves the classification of Supreme Court opinions.

No test reported here found that it does. The three generic stoplists in common use fell below the no-removal baseline in all twelve generic-list tests, on both tasks and under both n-gram ranges (Section 4). The task-specific lists, built by scoring the removal of each of roughly 18,500 candidate words and keeping the words whose removal helped by more than measurement error, moved held-out F1 by +0.0023 (95% CI [−0.0124, +0.0170]) on ideology, where a baseline F1 near 0.68 leaves room for a stoplist to help, and by +0.0001 ([−0.0082, +0.0085]) on law type, where a baseline near 0.92 leaves little. Both intervals include zero. Their edge over generic lists on the point estimates is not detected either; every one of the twelve paired intervals includes zero (Section 5.3). Most of the gap between the gains measured one word at a time and the gain from removing the words together traces to a shared pool of borderline opinions whose classifications many of the selected words shift in common, so that the single-word gains count the same opinions again and again (Section 5.3).

Those results support a claim narrower than that removal cannot help: No published stoplist has been validated for judicial text; three lists in wide use fell below the no-removal baseline in every test; and a systematic attempt to build a better one, which tested every candidate word and consumed more than four hundred hours of computation, produced no improvement on held-out opinions that the design could detect. Whether some other list could help these classifiers, however, remains open. An effect smaller than roughly ±0.015 F1 on ideology or ±0.008 on law type would have gone undetected on holdouts of this size (Section 6.3), and no search over combinations of words was run. A researcher who reaches for a stoplist in this setting nonetheless follows a convention with no demonstrated justification.

For classifying Supreme Court opinions with a TF-IDF pipeline, the right default is therefore to remove no stopwords at all. The recommendation rests on the costs and prospective benefits of each choice rather than on a claim that removal can never help (Section 6.1). Leaving stopwords in place forgoes at most a benefit the holdout evaluation could not detect. Removing them adds a preprocessing step and a discretionary choice among lists and thresholds. And in interpretable legal text-as-data, where the weighted feature is often itself the finding, removal reshapes the features a study is meant to interpret. The same reasoning extends to any preprocessing step decided word by word—stemming, lemmatization, and minimum-document-frequency cutoffs among them—and the same ablation test is available for each (Section 6.2). The burden of demonstrating a benefit now rests with anyone who would advocate stopword removal.

## Acknowledgments

AI tools (Claude and ChatGPT) were used to review and edit code, review literature, collect relevant sources, and edit and proofread the manuscript text. No AI tool is credited as an author, and the author is responsible for all content.

## ORCID iD

Gregory M. Dickinson https://orcid.org/0009-0006-8992-3929

## Ethical Considerations

Not applicable. The study uses only public court records and coded data and did not involve human participants.

## Author Contributions

Gregory M. Dickinson is the sole author and is responsible for the conception, design, analysis, and writing of the study.

## Funding

The author received no financial support for the research, authorship, and/or publication of this article.

## Declaration of Conflicting Interests

The author declared no potential conflicts of interest with respect to the research, authorship, and/or publication of this article.

## Data Availability Statement

The replication package—the SCDB-to-CAP matching code, the complete per-word ablation results for both tasks, and the analysis and evaluation scripts—will be deposited in a public repository with a persistent DOI. The source data are the Supreme Court Database (Spaeth et al., 2025) and the Caselaw Access Project (2024), both publicly available.

## Supplemental Material

Supplemental material for this article is available online.